\documentclass[10pt,twocolumn,letterpaper]{article}

\usepackage[pagenumbers]{cvpr} 

\usepackage{times}
\usepackage{soul}
\usepackage{url}
\usepackage[utf8]{inputenc}
\usepackage{graphicx}
\usepackage{svg}
\usepackage{amsmath}
\usepackage{amssymb}
\usepackage{amsthm}
\usepackage{booktabs}
\usepackage{comment}
\usepackage{algorithm}
\usepackage{algpseudocode}
\usepackage{pifont}   
\usepackage[switch]{lineno}
\definecolor{cvprblue}{rgb}{0.21,0.49,0.74}
\definecolor{predcolor}{HTML}{706AA1}
\usepackage[pagebackref,breaklinks,colorlinks,allcolors=cvprblue]{hyperref}

\usepackage{graphicx}
\def\paperID{11} 
\def\confName{CVPR}
\def\confYear{2026}

\title{Recall to Predict: Grounding Motion Forecasting in Interpretable Motion Bank}

\author{Abhishek Vivekanandan$^{1,2}$ \quad Ahmed Abouelazm$^{1}$ \quad J.~Marius Z\"ollner$^{1,2}$\\
$^{1}$FZI Forschungszentrum Informatik\\
$^{2}$Karlsruhe Institute of Technology (KIT)\\
Karlsruhe, Germany\\
{\tt\small \{vivekana\}@fzi.de}
}

\begin{document}
\maketitle
\begin{abstract}
Motion forecasting often requires trading interpretability for predictive accuracy. Standard anchor-based architectures rely on opaque latent queries that are highly prone to latent collapse, or naive trajectory sampling that limits multi-modal diversity. We propose an end-to-end differentiable framework that grounds predictions in a comprehensive "motion bank", a structured embedding space of physically realizable trajectories constructed via contrastive learning. Rather than regressing paths from a blank slate, our architecture dynamically retrieves explicit motion priors using a novel Anchor Retrieval Layer. This module adapts orthogonally initialized queries via a Dual-Level Gated Cross-Attention mechanism and executes discrete trajectory selection using a Straight-Through Gumbel-Softmax estimator to preserve continuous gradient flow. The retrieved semantically grounded anchors are then geometrically refined by a DETR-style decoder, optimized jointly with a Winner-Takes-All (WTA) kinematic Gaussian Mixture Model (GMM), a latent diversity penalty, and a soft-min weighted endpoint loss. By strictly conditioning the decoding phase on diverse, interpretable motion primitives, our approach eliminates the "black box" of standard latent queries while achieving competitive multi-modal accuracy on the Argoverse 2 and Waymo Open Motion datasets. Code is available at: \url{https://github.com/abviv/recall2predict} 
\end{abstract}

\section{Introduction}

In autonomous driving, motion forecasting models must generate future trajectories that are not only accurate but also kinematically feasible and physically compliant with the environment. To achieve this, trajectory sets are a natural choice for structural priors, as they directly encode the physically consistent behaviors observed in real-world driving data. Unlike models that enforce rigid heuristics or rely on abstract latent codes, trajectory-based approaches \cite{phan-minh_covernet_2020, biktairov_prank_2020} inherit realistic patterns directly from demonstrations. However, the central challenge in modern forecasting is not merely representing these trajectories, but determining how to integrate them as structured priors without sacrificing multi-modal diversity or model interpretability.

Historically, anchor-based methods have been widely adopted to inject strong directional intent and enable multi-modal predictions. However, the field's heavy optimization for endpoint-centric metrics (such as minFDE) has inadvertently sparked a race that sacrifices interpretability during the intermediate prediction steps. Learning anchors directly in the latent space~\cite{nayakanti_wayformer_2022,zhao_tnt_2021,wang_prophnet_2023,shi_motion_2023} yields compact but fundamentally opaque representations. Conversely, sampling fixed subsets of trajectories from the training distribution~\cite{varadarajan_multipath_2021,vivekanandan_ki-pmf_2024} tends to over-represent common behaviors while missing rare, safety-critical edge cases. Within this work, we argue that anchors should not be treated merely as abstract initialization points to minimize endpoint error, but instead they should serve as explicit, understandable priors that allow us to inspect the model's reasoning \textit{before} the decoding phase.

To realize this, we propose a fully differentiable, end-to-end forecasting framework that grounds its reasoning in a comprehensive, pre-trained motion bank. Leveraging the contrastively trained embeddings from ~\cite{vivekanandan_contrast_2025}, we establish a one-to-one mapping between physically realizable future trajectories and semantically structured latent vectors. Rather than regressing paths from a blank slate, our architecture dynamically queries this bank using a context-aware Anchor Retrieval Layer. By utilizing a Straight-Through Gumbel-Softmax estimator, the network performs discrete, attention-based retrieval of exact trajectory priors while maintaining continuous gradient flow. This explicit retrieval mechanism provides interpretability, revealing exactly which motion primitives the network considers prior to the black-box decoding phase. Finally, a DETR-style decoder \cite{carion_detr_2020} fuses these semantic priors with spatial scene context to regress continuous kinematic states and geometric offsets, smoothly bridging the gap between discrete intent and continuous motion.

\textbf{Our main contributions are as follows:}
\begin{itemize}
    \item We introduce an end-to-end trajectory forecasting framework that leverages a pre-trained motion bank to inject explicit, physically realizable motion priors into the decoding process, enabling the accurate retrieval of context-relevant driving behaviors.
    \item We propose a differentiable, attention-based Anchor Retrieval Layer. By decoupling the forward discrete selection from the backward continuous gradients via a Straight-Through Estimator (STE), we establish highly interpretable intermediate representations that expose the model's decision-making process in the early layers.
    \item We design a multi-objective refinement strategy, combining a Winner-Takes-All kinematic Gaussian Mixture Model (GMM), offset regularization, and latent diversity constraints -- that geometrically fine-tunes the selected anchors while strictly preserving their foundational semantic intent.
\end{itemize}

\section{Related Work}

\paragraph{Map-Conditioned and Trajectory Set Methods.} Goal-conditioned approaches (TNT \cite{zhao_tnt_2021}, DenseTNT \cite{gu2021densetnt}) and Frenet-based formulations (PRIME~\cite{song2022learning}) improve multimodal diversity but rely heavily on heuristic endpoint generation and high-definition (HD) map traversals. In real-world deployments, this strict map dependence propagates perception noise directly into the motion prior. To reduce map reliance, prior works~\cite{vivekanandan_ki-pmf_2024, vivekanandan_efficient_2024} explored sampling predefined trajectory sets from the training distribution. However, naive geometric sampling struggles to balance common patterns with rare, safety-critical behaviors without introducing significant bias.

\paragraph{Transformer Queries and Abstract Anchors.} Recent architectures generate diverse futures via learnable proposal queries. DeMo~\cite{zhang2024decoupling} decouples directional and dynamic state queries, while MTR~\cite{shi_mtr_2024} and ControlMTR~\cite{sun_controlmtr_2024} anchor predictions using clustered dataset \textit{endpoints} combined with map-heavy drivable constraints. While effective, anchoring on isolated endpoints or abstract embeddings, rather than full trajectory profiles, weakens interpretability and physical feasibility. Furthermore, maintaining query diversity remains an open challenge. Orthogonal initialization~\cite{wang_prophnet_2023} often succumbs to attention collapse, where gradients pull multiple queries toward the same dominant historical tokens coming in from a single context source. On the anchor side, EDA~\cite{lin_eda_2023} proposes to move from static priors to evolving spatial anchors and by applying Non-Maximum Suppression (NMS). Yet, EDA’s anchors remain opaque latent vectors updated via continuous regression, lacking grounding in a globally valid physical space.

\paragraph{Our Approach.} We fundamentally depart from the abstract queries of DeMo and EDA, and the endpoint-centric priors of MTR. To eliminate brittle map dependencies and abstract initializations, we introduce an end-to-end differentiable Anchor Retrieval Layer. Rather than regressing paths from spatial heuristics, our model seeds orthogonally initialized queries solely from observed kinematics and social interactions. By utilizing a STE to query a pre-trained motion bank, we retrieve \textit{full, physically realizable trajectory states} as an intermediate representation. This provides a transparent, kinematically valid initialization that actively prevents latent collapse in intermediate layers, bridging the gap between explicit trajectory sets and continuous state refinement.

\begin{figure*}[t]
    \centering
    \includegraphics[width=0.9\textwidth, keepaspectratio]{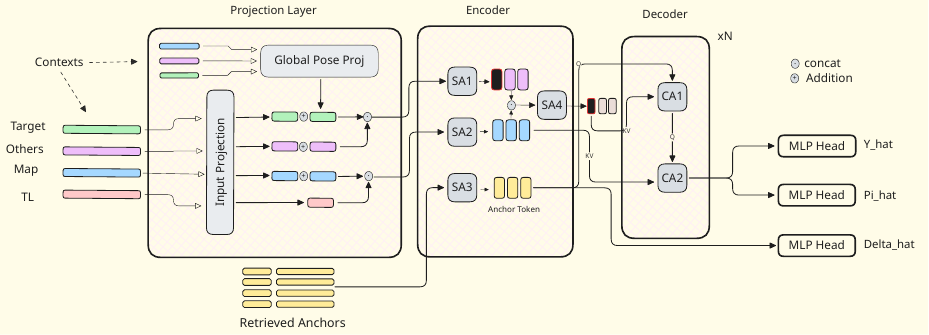}
    \caption{\textbf{Overview of the Proposed Architecture.} The system initially processes heterogeneous driving context using a PointNet-based projection layer. Because agent-centric pooling inherently strips absolute geometric grounding, global pose features are explicitly restored to the embeddings via residual additions. Guided by the projected context, an auxiliary retrieval module fetches $N_{q}$ discrete Anchor Tokens from a pre-trained motion bank. The scene features are then deeply contextualized by a Factorized Scene Encoder, which employs modality-specific self-attention (SA) followed by joint global fusion. In the final stage, a DETR-style decoder iteratively refines the anchor tokens via cross-attention (CA) against the fused scene representation. The refined queries are processed by parallel MLP heads to predict continuous kinematic states, mode confidence scores, and anchor endpoint offsets.}
    \label{fig:overall_arch}
\end{figure*}
\section{Model Architecture}

We formulate our model as a fully differentiable, end-to-end trainable Encoder-Decoder architecture. A key component of this system is the Anchor Retrieval Layer, which functions as an auxiliary encoder to retrieve candidate anchors from a pre-trained motion bank. Specifically, the model derives representations from the contexts to achieve parity with the pre-trained embedding space. The discrete anchor selection process is then approximated via a soft Gumbel-Softmax relaxation \cite{jang2017categoricalreparameterizationgumbelsoftmax}, thereby preserving the differentiability of the entire pipeline.

\subsection{Pre-trained Embedding Motion Bank}

At its core, our framework leverages a diverse library of prototypical movements to guide trajectory forecasting. Instead of forcing the decoder to regress future paths from scratch, our model queries a pre-constructed "motion bank" to fetch candidate trajectories that closely align with the observed environment.

To efficiently represent and match these motion patterns, we utilize the publicly available pre-trained embeddings from~\cite{vivekanandan_contrast_2025}. Derived from a contrastive framework that maps geometrically similar trajectories to proximal points in the latent space, these embeddings provide a highly structured feature space for identifying useful motion priors. The resulting motion bank is formulated as a collection of key-value pairs, directly linking each trajectory (referred to synonymously as an "anchor" or a "path") to its respective embedding. 

\subsection{Input Representation and Projection}
\label{sec:input_projection}

Our framework adopts an agent-centric coordinate system, taking four primary modalities as input: the target agent's history $\mathcal{A}_{\text{hist}} \in \mathbb{R}^{T_s \times D_a}$, the historical states of neighboring actors $\mathcal{N}_{\text{hist}} \in \mathbb{R}^{N_a \times T_s \times D_a}$, the static lane polylines $\mathcal{M}_{\text{poly}} \in \mathbb{R}^{N_m \times P \times D_m}$, and the traffic light states $\mathcal{TL}_{\text{state}} \in \mathbb{R}^{N_{tl} \times T_s \times D_{tl}}$. Here, $T_s$ denotes the number of past timesteps, $N_a$ is the number of surrounding actors, $N_m$ is the number of lane segments, $P$ is the number of waypoint nodes per polyline, and $N_{tl}$ represents the number of traffic lights. The dimensions $D_a$, $D_m$, and $D_{tl}$ denote the modality-specific attribute features, which inherently encapsulate the 2D spatial $(x, y)$ coordinates alongside auxiliary states such as heading and velocity.

To establish a unified dimensionality across these heterogeneous inputs, we process each modality through a PointNet-style \cite{Qi_2017_CVPR} encoder following \cite{gao_vectornet_2020}. This module iteratively interleaves per-timestep (or per-node) Multi-Layer Perceptrons (MLPs) with global max-pooling. At each layer, the globally pooled feature vector is concatenated back to the sequence, enabling the network to capture both local temporal dynamics and global trajectory statistics. 

Crucially, we employ a dual-resolution projection strategy to serve different downstream architectural needs. For the Encoder, the final representation of each element is obtained by max-pooling across the temporal or spatial dimension, effectively collapsing the sequence into a single, fixed-dimensional embedding per agent or polyline. Conversely, for the Anchor Retrieval Layer, we bypass this final temporal pooling for the target agent, preserving its dense, per-timestep representations ($\mathbb{R}^{T_s \times D}$). This explicit retention of temporal granularity is what enables the orthogonal latent queries to attend to distinct, fine-grained historical moments prior to anchor selection.

\paragraph{Explicit Pose Reprojection.}
While global max-pooling is highly efficient for dimensionality reduction, it inherently destroys the absolute spatial layout of the polylines within the agent-centric coordinate frame. Pairwise relative representations, such as those used in HPTR \cite{zhang_real-time_2023}, naturally bypass this issue; however, in an agent-centric model, the loss of geometric grounding early in the network degrades spatial reasoning. 

To resolve this, we introduce an explicit pose reprojection mechanism. We extract the geometric state of each element; specifically, the final timestep's $[x, y, heading\_angle]$ for agents, and the normalized centroid and direction for lane polylines. These geometric features are projected through a two-layer MLP to match the hidden dimension $D$, and are subsequently injected into the max-pooled features via element-wise addition. This operation acts as a continuous spatial positional encoding, restoring the crucial relative geometric layout without sacrificing the computational efficiency of the PointNet encoder.

\subsection{Anchor Retrieval Layer}
\label{sec:anchor_selection}
\begin{figure}[t]
    \centering
    \includegraphics[width=1.0\columnwidth, keepaspectratio]{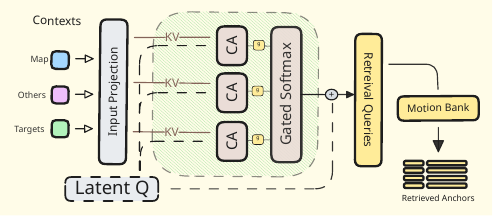}
    \caption{Anchor Retrieval Layer. Orthogonally initialized latent queries $\mathbf{Q}_{\text{base}}$ distill context (Map features are shown for ablation purposes only and are excluded from the main model) representations via parallel cross-attention (CA) blocks. These are balanced by a gated softmax and residually combined to fetch discrete trajectory anchors from the pre-trained Motion Bank.}
    \label{fig:Anchor Retrieval}
\end{figure}

To provide structural priors and an intermediate interpretable representation, our architecture features a fully differentiable Anchor Retrieval Layer. This module acts as an auxiliary encoder that maps provided contexts to the latent space of the pre-trained motion bank, outputting $N_{q}$ (defaults to 6 unless stated otherwise) distinct anchor proposals (which represent future states) and their corresponding embeddings to seed the downstream decoder.

\paragraph{Context-Aware Query Adaptation.}
We initialize $N_{q}$ learnable, orthogonal latent queries $\mathbf{Q}_{\text{base}} \in \mathbb{R}^{N_{q} \times D}$ which learn from the heterogeneous contexts of the target agent and the surrounding scene to extract representations.  To prevent any single modality from dominating the latent representation, we propose a \textbf{Dual-Level Gated Cross-Attention} mechanism. 

Instead of cross-attending to a concatenated sequence of projected contexts, $\mathbf{Q}_{\text{base}}$ attends to them through independent, modality-specific cross-attention pathways, each producing an attended representation $\mathbf{H}_m \in \mathbb{R}^{N_q \times D}$ for modality $m$. The adaptation relies on a two-tiered gating strategy:

\begin{enumerate}

\item \textbf{Micro-Level Gating:} A query-specific sigmoid gate ($g_m$) evaluates the concatenation of the base query and $\mathbf{H}_m$, allowing each query to determine how much context to absorb. Indicated as \textit{g} in \cref{fig:Anchor Retrieval}, sitting between the CA block and the Gated Softmax block.

\item \textbf{Macro-Level Routing:} A global softmax computes importance weights ($w_m$) to balance relevance across modalities. Crucially, this softmax includes a learnable ``null'' sink, which safely absorbs probability mass when the surrounding context is uninformative.
\end{enumerate}

The final adapted queries are formed via a weighted residual update:
\begin{equation}
\mathbf{Q}_{\text{adapt}} = \mathbf{Q}_{\text{base}} + \sum_{m \in \{\mathcal{A}_{\text{hist}}, \mathcal{N}_{\text{hist}}, 
\mathcal{M}_{\text{poly}}\}} w_m \, (g_m \odot \mathbf{H}_m)
\end{equation}
Note that $\mathcal{TL}_{\text{state}}$ is intentionally excluded from the retrieval queries; traffic light context is instead fused downstream within the scene encoder (\cref{sec:scene_encoder}). This fusion strategy ensures that the queries are dynamically enriched by the contexts while strictly preserving their orthogonal diversity.

\paragraph{Past-to-Future Bridging via Cosine Attention.}
The pre-trained embeddings in the motion bank exclusively represent the \textit{future} geometries of trajectories, whereas our adapted queries encode the \textit{historical} and contextual state of the scene. To bridge these temporal domains, we compute the cosine similarity between the $\ell_2$-normalized adapted queries $\hat{\mathbf{Q}}_{\text{adapt}}$ and the normalized-Frozen bank embeddings $\mathbf{E} \in \mathbb{R}^{B \times D_{\text{emb}}}$, where $B$ denotes the total number of trajectories in the motion bank and $D_{\text{emb}}$ their embedding dimension:
\begin{equation}
    \mathbf{Z} = \hat{\mathbf{Q}}_{\text{adapt}} \mathbf{E}^\top
\label{eq:cosine_sim}
\end{equation}
This learnable mapping forces the network to align its historical context representation with the geometric intent captured in the future trajectory space.

\paragraph{Differentiable Selection via Straight-Through Gumbel-Softmax.}
To ensure end-to-end differentiability during the discrete retrieval of $N_{q}$ trajectories, we employ a STE. Given attention logits $\mathbf{Z} \in \mathbb{R}^{N_q \times B}$ over the motion bank, we compute a continuous distribution:
\begin{equation}
    \pi = \text{Softmax}(\mathbf{Z} / \tau)
\label{eqn:temp_anneal}
\end{equation}
where $\tau$ is annealed over training to transition from exploration to exploitation \cite{radford2021learningtransferablevisualmodels}. To decouple the forward and backward passes, we generate a hard one-hot selector $\mathbf{Y}_{\text{hard}} \in \{0,1\}^{N_q \times B}$ via an $\arg\max$ operation, and apply the straight-through trick:
\begin{equation}
    \mathbf{Y}_{\text{ST}} = \mathbf{Y}_{\text{hard}} + \big(\pi - \text{sg}(\pi)\big)
\end{equation}
where $\text{sg}(\cdot)$ is the stop-gradient operator. We then extract features from the pre-trained motion bank $\mathbf{E} \in \mathbb{R}^{B \times D_{\text{emb}}}$ via the matrix multiplication $\mathbf{E}_{\text{ret}} = \mathbf{Y}_{\text{ST}} \mathbf{E}$. Because $\mathbf{Y}_{\text{ST}}$ acts strictly as a one-hot mask in the forward pass, this operation cleanly retrieves $N_q$ exact, unblended embeddings. Finally, these embeddings are projected to the hidden dimension $D$ via an MLP, and residually summed with the adapted queries and projected trajectory coordinates to form the dense \textit{Anchor Tokens} $\mathbf{A} \in \mathbb{R}^{N_q \times D}$ that seed the decoder as a Query.

\subsection{Factorized Scene Encoder}
\label{sec:scene_encoder}

To capture the complex structural dynamics of the driving environment before decoding, we employ a factorized-then-fused encoding strategy. This approach processes dynamic agents and static map elements independently, building robust domain-specific priors before executing cross-modal reasoning. The encoder operates in three stages:

\begin{enumerate}
    \item \textbf{Intra-Agent Social Encoding:} We apply a stack of self-attention blocks exclusively over the concatenated agent tokens ($\mathcal{A}_{\text{hist}}$ and $\mathcal{N}_{\text{hist}}$). This captures critical scene-level dynamic interactions, such as yielding and car-following behaviors.
    \item \textbf{Intra-Environment Topological Encoding:} In parallel, a separate self-attention stack processes the $\mathcal{M}_{\text{poly}}$ concatenated with $\mathcal{TL}_{\text{state}}$. By integrating traffic light features directly into the lane stream, the network models both the static structural relationships and the regulatory constraints of the road graph.
    \item \textbf{Joint Modality Fusion \& Focal Extraction:} The updated agent and environment streams are concatenated into a unified sequence and processed via joint self-attention, facilitating cross-domain message passing. Following this global fusion, the contextualized environmental tokens (lanes and traffic lights) are collected as $\mathbf{M}_{\text{env}}$, and we extract \textit{only} the token corresponding to the target agent ($\mathbf{F}_{\text{target}}$). Analogous to a \texttt{[CLS]} token in Vision Transformers \cite{dosovitskiy2021imageworth16x16words}, this \textbf{focal agent token} (black shaded box from \cref{fig:overall_arch}) now encapsulates the deeply contextualized representation of the entire scene, and is the sole dynamic state forwarded to the decoder. Meanwhile, the contextualized tokens for neighboring agents are routed to a separate dense predictor for single-shot trajectory regression.
\end{enumerate}
Structuring the encoder to learn modality-specific representations prior to global fusion prevents optimization instability, while the focal extraction drastically reduces the computational burden of the downstream decoder.

\subsection{Iterative Refinement Decoder}
\label{sec:decoder}

To generate the final predictions, we employ an iterative refinement decoder. Unlike standard DETR-style architectures \cite{carion_detr_2020} that rely on learnable latent queries initialized from scratch, our decoder directly utilizes the \textit{Anchor Tokens} (PGQA ~\cref{alg:pagqa} when $N_{q}$ $>$ 6) as its initial queries. Before entering the decoder, these anchor tokens undergo an initial self-attention phase to model inter-modal relationships among the proposed anchors. This ``anchors-as-queries'' formulation ensures a strict 1-to-1 correspondence between the latent queries and the retrieved physical trajectories, preserving their geometric interpretability throughout the network.

Within each decoder layer, the queries are updated through a sequential Cross-Attention (CA) mechanism to aggregate scene context. Let $\mathbf{Q}^{(l-1)}$ denote the input queries to layer $l$. The refinement proceeds by attending to two distinct contextual sources (acting as Keys and Values):

\begin{enumerate}
    \item \textbf{Target Context:} The queries first attend to the focal agent token, extracting the target's specific, scene-aware dynamic history:
    \begin{equation}
        \mathbf{Q}^{(l)}_1 = \text{CA}\left(\text{Q}=\mathbf{Q}^{(l-1)},\; \text{KV}=\mathbf{F}_{\text{target}}\right)
    \end{equation}
    
    \item \textbf{Environment Context:} The queries then attend to the refined environmental tokens ($\mathbf{M}_{\text{env}}$), which encompass both lane geometry and traffic light states, to incorporate structural constraints and road topology:
    \begin{equation}
        \mathbf{Q}^{(l)} = \text{CA}\left(\text{Q}=\mathbf{Q}^{(l)}_1,\; \text{KV}=\mathbf{M}_{\text{env}}\right)
    \end{equation}
\end{enumerate}
Because the queries are already grounded in the retrieved anchors, no further cross-attention to the anchor bank is required.

\paragraph{Prediction Heads.}
The refined queries from the final decoder layer, $\mathbf{Q}^{(L)} \in \mathbb{R}^{K \times D}$, and the pre-decoder \textit{Anchor Tokens}, $\mathbf{Q}^{(0)} \in \mathbb{R}^{K \times D}$, are processed by parallel MLP heads:
\begin{align}
    \hat{\mathbf{Y}} &= \mathrm{MLP}_{\text{kin}}(\mathbf{Q}^{(L)}) \in \mathbb{R}^{K \times T_f \times D_{\text{kin}}}, \\
    \hat{\boldsymbol{\pi}} &= \mathrm{MLP}_{\text{conf}}(\mathbf{Q}^{(L)}) \in \mathbb{R}^{K \times 1}, \\
    \Delta\hat{\mathbf{p}} &= \mathrm{MLP}_{\text{off}}(\mathbf{Q}^{(0)}) \in \mathbb{R}^{K \times 2}.
\end{align}
Here, $T_f$ denotes the number of predicted future timesteps, $D_{\text{kin}}$ is the per-step kinematic state dimension (spatial means, covariance parameters, velocity, and heading), $\hat{\mathbf{Y}}$ regresses the continuous future state sequence for the $K$ representative modes, and $\hat{\boldsymbol{\pi}}$ yields the confidence distribution over them. 

Crucially, the offset head derives continuous 2D spatial residuals ($\Delta\hat{\mathbf{p}}$) directly from the pre-decoder \textit{Anchor Tokens} rather than the refined queries $\mathbf{Q}^{(L)}$. This strict decoupling prevents the decoder from lazily compensating for poor anchor selections using its deep contextual refinement. By predicting offsets from the pre-decoder anchor embeddings, the network geometrically bridges the discretization gap of the retrieval bank during training, and these offsets can be safely discarded at inference.
\subsection{Learning Objectives}
\label{sec:loss_functions}

Our model is trained end-to-end using a composite objective function that balances accurate multi-modal trajectory regression with the retrieval of geometrically diverse and relevant priors:
\begin{equation}
    \mathcal{L}_{\text{total}} = \lambda_{\text{motion}}\,\mathcal{L}_{\text{motion}} + 0.1\,\mathcal{L}_{\text{endpoint}} + \mathcal{L}_{\text{div}}
\end{equation}

\subsubsection{Winner-Takes-All Kinematic Loss}
We adopt a Winner-Takes-All (WTA) strategy to avoid mode-averaging: only the mode $k^*$ with the lowest cumulative displacement across all future timesteps is penalized. Its spatial error is minimized via the Negative Log-Likelihood of a bivariate Gaussian, where the network jointly predicts the 2D mean and covariance parameters. This is augmented with a Huber loss on velocity ($\mathcal{L}_{\text{vel}}$), a cosine penalty on heading ($\mathcal{L}_{\text{yaw}}$), and a cross-entropy term that sharpens the confidence distribution toward $k^*$:
\begin{equation}
    \mathcal{L}_{\text{motion}} = w_{\text{pos}}\,\mathcal{L}_{\text{NLL}} + w_{\text{vel}}\,\mathcal{L}_{\text{vel}} + w_{\text{yaw}}\,\mathcal{L}_{\text{yaw}} + w_{\text{conf}}\,\mathcal{L}_{\text{CE}}(\hat{\boldsymbol{\pi}}, k^*)
\end{equation}
where $w_{\text{pos}}$, $w_{\text{vel}}$, $w_{\text{yaw}}$, and $w_{\text{conf}}$ are per-agent-type scalar weights that allow independent tuning across vehicles, pedestrians, and cyclists.

\paragraph{Dual-Objective Endpoint Loss.} 
This loss evaluates the refined anchor endpoints, $\hat{\mathbf{p}}_n = \mathbf{p}_n^{\text{anch}} + \Delta\hat{\mathbf{p}}_n$, where $\Delta\hat{\mathbf{p}}_n$ is the continuous 2D spatial correction predicted by the offset head for the $n$-th anchor. By combining the discrete base anchor with continuous residuals, this formulation serves as a dual objective, simultaneously guiding the discrete anchor selection and optimizing local geometric accuracy. We apply a soft-min weighting over the $N_q$ anchor proposals:
\begin{equation}
    \mathcal{L}_{\text{endpoint}} = \sum_{n=1}^{N_q} w_n \Big[ \text{Huber}(\hat{\mathbf{p}}_n, \mathbf{p}^*) \Big]
\end{equation}
where $\mathbf{p}^*$ is the ground truth endpoint and $w_n$ represents the soft-min weight derived from the distance between $\hat{\mathbf{p}}_n$ and $\mathbf{p}^*$. Coupling the offset directly to the pre-trained anchors ensures the network relies on the retrieved motion primitives for global trajectory structure, utilizing the continuous offsets solely for local calibration.

\begin{table*}[t]
    \centering
    \resizebox{0.85\textwidth}{!}{%
        \begin{tabular}{lcccccc}
            \toprule
            Model & minADE$_6\downarrow$ & minADE$_1\downarrow$ & minFDE$_6\downarrow$ & minFDE$_1\downarrow$ & MissRate$\downarrow$ & brier-minFDE$\downarrow$ \\
            \midrule
            ForecastMAE (Scratch) \cite{cheng2023forecast}              & 0.73 & 1.84   & 1.42 & 4.60   & 0.19 & 2.06 \\
            SIMPL \cite{zhang2024simplsimpleefficientmultiagent}                         & 0.72 & 2.03 & 1.42 & 5.50 & 0.19 & 2.05 \\
            HPTR \cite{zhang_real-time_2023}                         & 0.73 & --   & 1.43 & --   & 0.19 & 2.03 \\
            GoReLa \cite{cui2022gorelarelativeviewpointinvariantmotion}                        & 0.76 & 1.82 & 1.48 & 4.62 & 0.22 & 2.01 \\
            MTR \cite{shi_motion_2023}                        & 0.73 & 1.74 & 1.44 & 4.39 & 0.15 & 1.98 \\
            EMP-D \cite{prutsch_efficient_2024}                         & 0.71 & 1.75 & 1.37 & 4.35 & 0.17 & 1.98 \\
            QCNET \cite{Qi_2017_CVPR}                         & 0.65 & 1.69 & 1.29 & 4.30 & 0.16 & 1.91 \\
            SmartRefine \cite{zhou2024smartrefinescenarioadaptiverefinementframework}     & 0.63 & 1.65 & 1.17 & 4.17 & 0.15 & 1.86 \\
            DeMo \cite{zhang2024demo}                          & \textbf{0.61} & \textbf{1.41} & \textbf{1.17} & \textbf{3.74} & \textbf{0.13} & \textbf{1.84} \\
            \midrule
            \textbf{R2P-M(Ours)}           & 0.69 & 1.75   & 1.33 &  4.37   & 0.17 & 2.03 \\
            
            \bottomrule
        \end{tabular}
    }
    \caption{Performance on Argoverse 2 Test dataset.}
    \label{tab:av2_test}
\end{table*}

\begin{table*}[t]
    \centering
        \resizebox{0.6\textwidth}{!}{%
            \begin{tabular}{lccccc}
                \toprule
                Model & minADE$_6\downarrow$ & minFDE$_6\downarrow$ & MissRate$_6\downarrow$ & mAP$\uparrow$ & OR$\downarrow$ \\
                \midrule
                MTRv3 (Ensemble)  & 0.5539 & 1.1041 & 0.1097 & 0.4877 & -- \\
                ControlMTR \cite{sun_controlmtr_2024}     & 0.5904 & 1.2010 & 0.1323 & 0.4220 & -- \\
                MTR \cite{shi_motion_2023}            & 0.6046 & 1.2251 & 0.1366 & 0.4160 & -- \\
                MTR++ \cite{shi_mtr_2024}          & 0.5912 & 1.1986 & 0.1296 & 0.4351 & -- \\
                HPTR \cite{zhang_real-time_2023}           &0.5378&1.0923 &0.1326&0.4150& -- \\
                EDA \cite{lin_eda_2023}            & 0.5708 & 1.1730 & 0.1178 & 0.4353 & 0.1273 \\
                \midrule
                \textbf{R2P-L (Ours)} & \textbf{0.5579} & \textbf{1.1247} & \textbf{0.1343} & \textbf{0.3371} & \textbf{0.1314} \\
                \textbf{R2P-M (Ours)} & \textbf{0.5474} & \textbf{1.1132} & \textbf{0.1387} & \textbf{0.2855} & \textbf{0.1418} \\
                \bottomrule
            \end{tabular}
        }
        \caption{Performance on WOMD Validation dataset.}
        \label{tab:waymo_val}
\end{table*}

\subsubsection{Latent Diversity Loss}
To prevent retrieval queries from collapsing, where the retrieval mechanism repeatedly fetches highly correlated anchors regardless of scene context---we enforce strict orthogonality in the latent space. Using the $\ell_2$-normalized adapted queries $\hat{\mathbf{Q}}_{\text{adapt}}$, we compute the cosine similarity matrix $\mathbf{S} = \hat{\mathbf{Q}}_{\text{adapt}} \hat{\mathbf{Q}}_{\text{adapt}}^\top$ and penalize its deviation from the identity matrix:
\begin{equation}
    \mathcal{L}_{\text{div}} = \lambda_{\text{div}} \| \mathbf{S} - \mathbf{I} \|_F^2
\end{equation}
where $\| \cdot \|_F$ denotes the Frobenius norm. This acts as a repulsive force, guaranteeing that the $N_{q}$ selected anchors encapsulate distinct semantic intents (e.g., yielding distinct left, straight, and right-turn priors).

\section{Experimental Setup}

\textbf{Datasets and Metrics.} We train and evaluate our models on the Waymo Open Motion Dataset (WOMD) \cite{Ettinger_2021_ICCV} and the Argoverse 2 (AV2) \cite{wilson2023argoverse2generationdatasets} dataset for the single-agent marginal motion prediction task. For evaluation, we report our results using the Brier minimum Final Displacement Error over six predictions (brier-minFDE) for AV2, and the mean Average Precision (mAP) for WOMD. While we conduct extensive ablation studies primarily on the AV2 validation dataset, we include results on WOMD to ensure the broader comparability and generalization of our approach.
\newline \\
\textbf{Implementation Details.} Our model processes a highly efficient, constrained context size of only 256 map polylines - significantly fewer than the 1024 used by HPTR \cite{zhang_real-time_2023} and 768 by MTR++ \cite{shi_mtr_2024}. Furthermore, we restrict the dynamic context to 32 surrounding actors and only the single traffic light state closest to the target agent. We employ a pre-trained motion bank containing 4096 trajectories constructed from the training distribution and use the same for evaluating test and validation baselines. Crucially, these priors possess an embedding dimension of 128, matching the decoder's hidden dimension ($D=128$). Following standard practice for marginal motion prediction, our network outputs $K=6$ future trajectory modes directly, entirely circumventing the need for Non-Maximum Suppression (NMS) or other computationally expensive post-processing heuristics.

\begin{figure*}[t]
    \centering
    \begin{subfigure}[b]{0.32\textwidth}
        \fbox{\includegraphics[width=\textwidth]{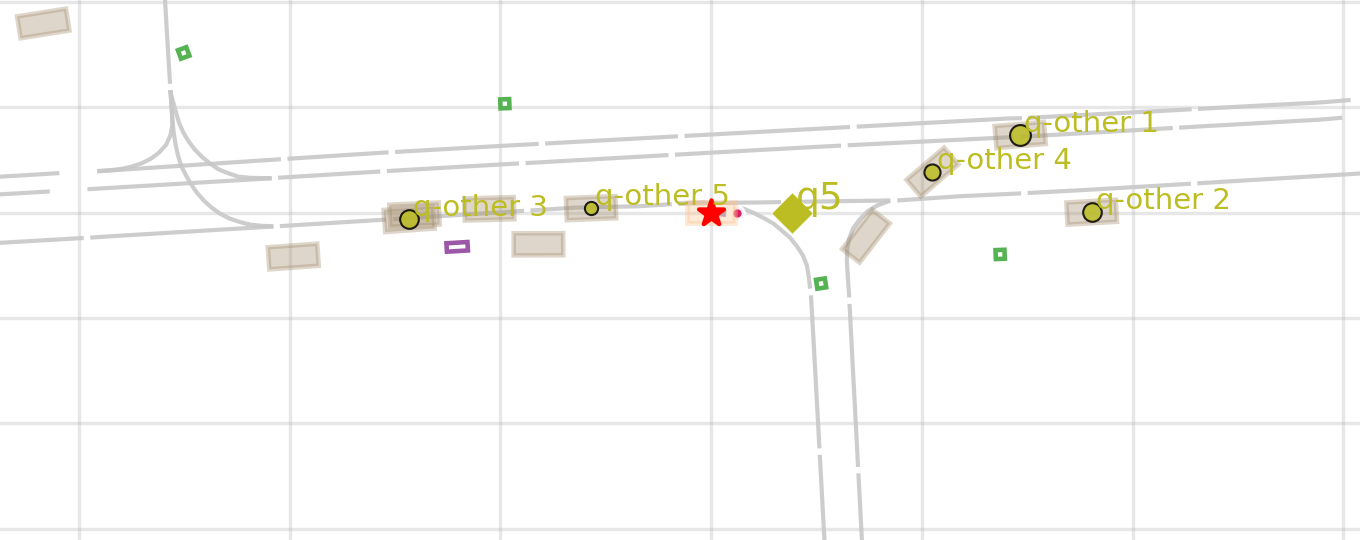}}
    \end{subfigure}
    \hfill
    \begin{subfigure}[b]{0.32\textwidth}
        \fbox{\includegraphics[width=\textwidth]{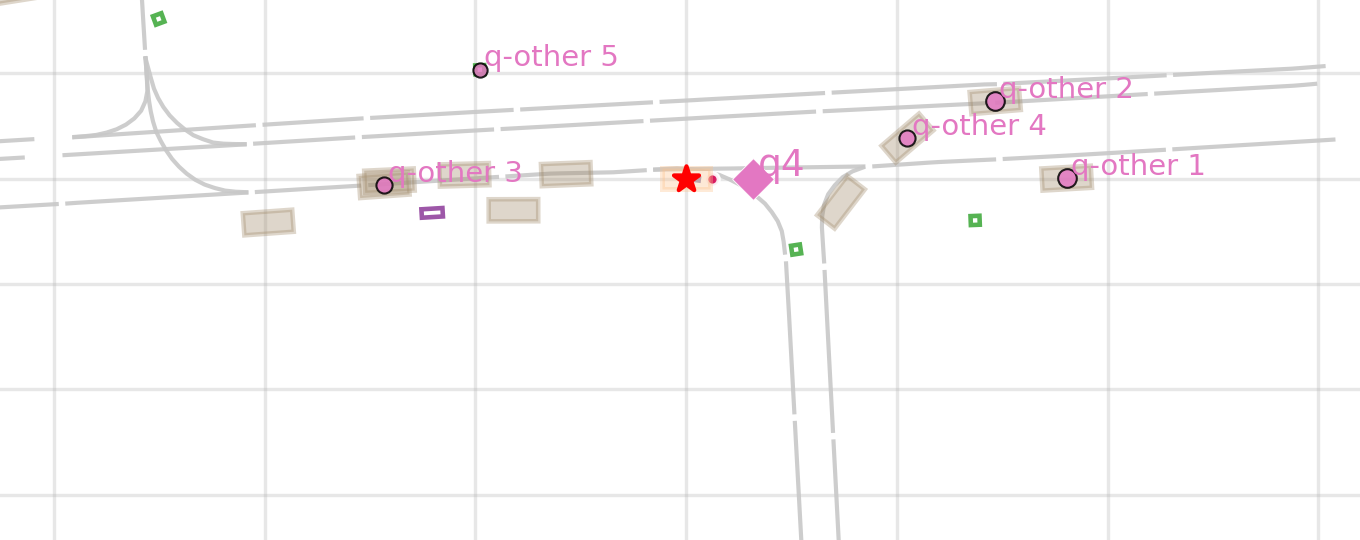}}
    \end{subfigure}
    \hfill
    \begin{subfigure}[b]{0.32\textwidth}
        \fbox{\includegraphics[width=\textwidth]{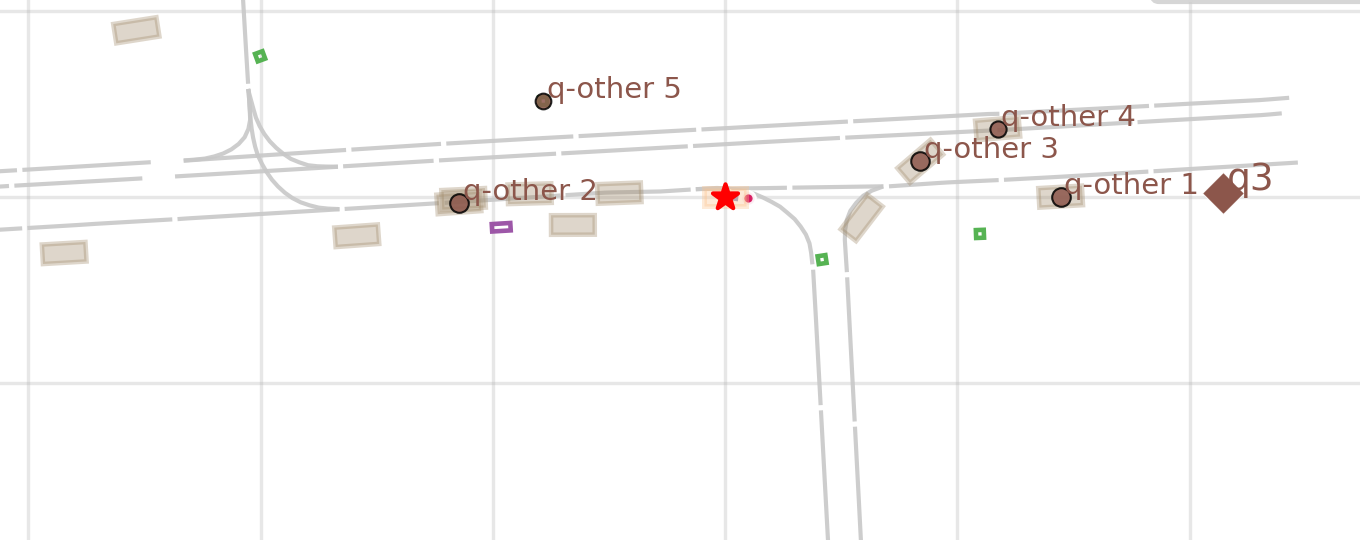}}
    \end{subfigure}

    \vspace{0.5em}

    \begin{subfigure}[b]{0.32\textwidth}
        \fbox{\includegraphics[width=\textwidth]{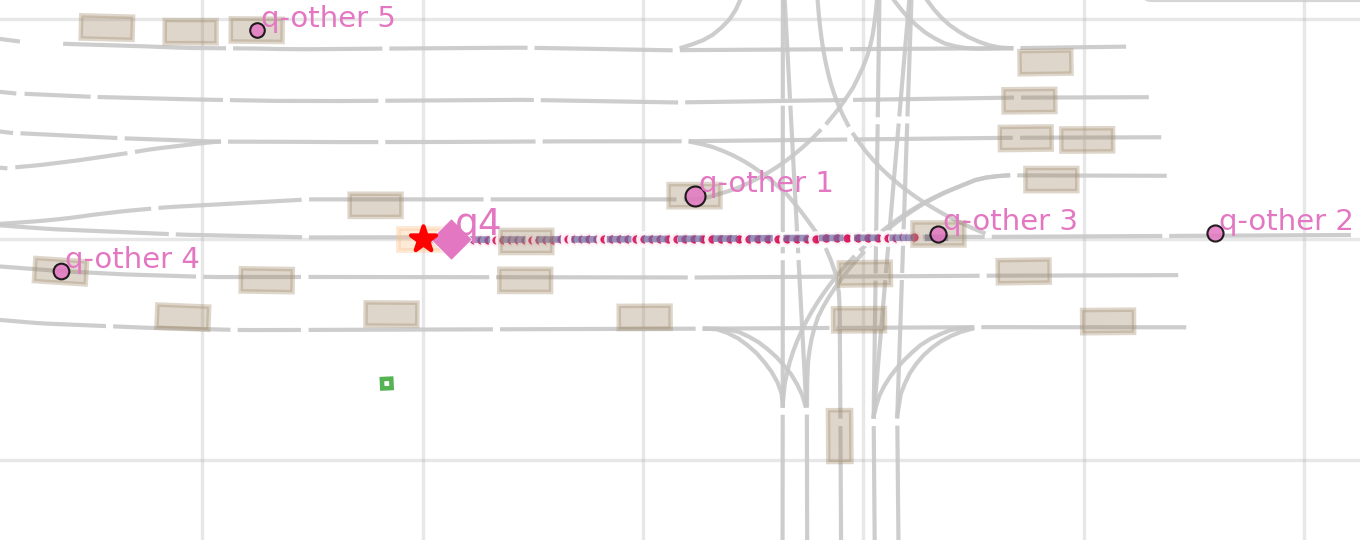}}
    \end{subfigure}
    \hfill
    \begin{subfigure}[b]{0.32\textwidth}
        \fbox{\includegraphics[width=\textwidth]{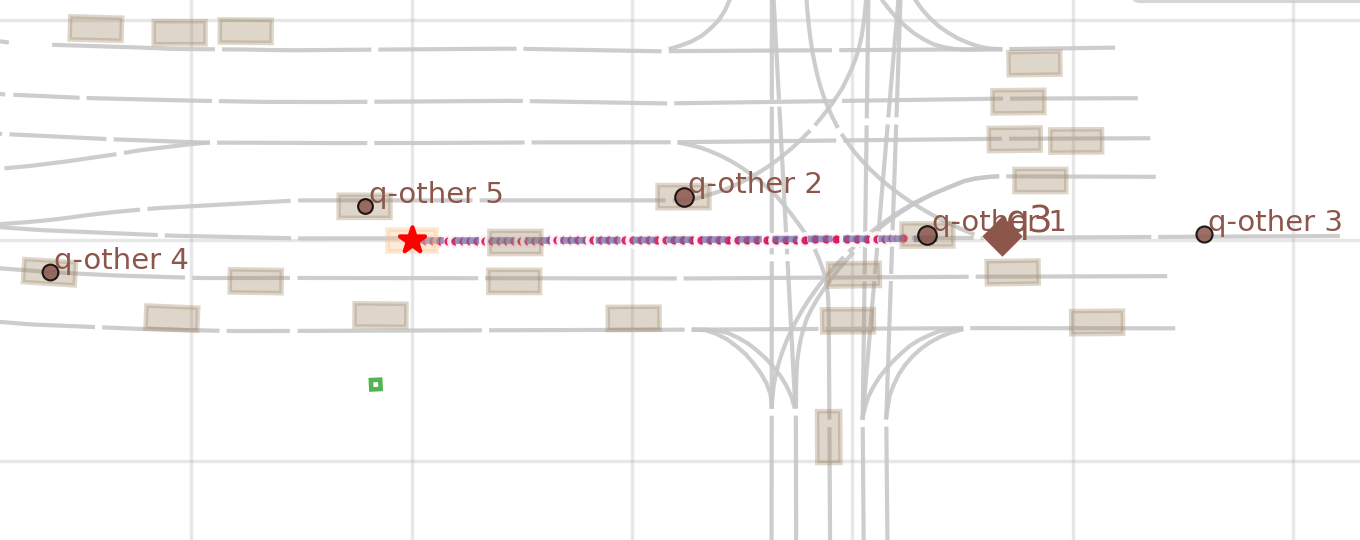}}
    \end{subfigure}
    \hfill
    \begin{subfigure}[b]{0.32\textwidth}
        \fbox{\includegraphics[width=\textwidth]{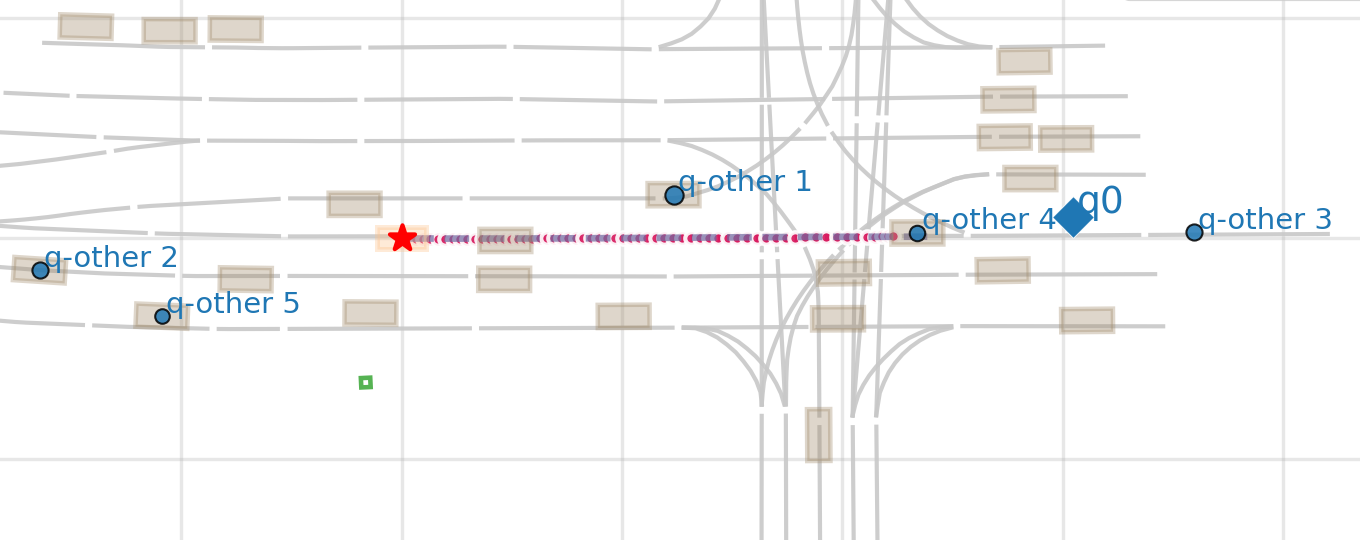}}
    \end{subfigure}

    \caption{Each row depicts a distinct driving scenario, illustrating three randomly selected latent queries alongside the top five scene elements they attend to (derived from cross-attention weights). This transparent retrieval mechanism highlights the inherent interpretability of our framework, allowing us to observe exactly how trajectory anchors are contextually molded by the environment prior to decoding. Notably, we observe that the queries predominantly aggregate features from neighboring actors, relying more heavily on social context than the target agent's own historical state to formulate their adapted representations.}
    \label{fig:4 interp_explanation}
\end{figure*}

\textbf{Training Details.}
To prevent latent collapse among the anchor proposals during training, we apply temperature annealing to \cref{eqn:temp_anneal}. Specifically, we employ a cosine annealing schedule that gradually reduces the temperature from an initial value of 5.0 to a final value of 0.25. For the AV2 dataset, we train the model for 75 epochs, achieving convergence in approximately 16 hours across 4 NVIDIA A100 (40GB) GPUs following DDP with a per-GPU batch size of 42. We utilize the OneCycleLR \cite{smith2018superconvergencefasttrainingneural} scheduling policy with a peak Learning Rate (LR) of $1.4 \times 10^{-3}$, an initial and a final division factor of 20 and 50, respectively, with AdamW optimizer having a weight decay of $1 \times 10^{-2}$. The LR schedule incorporates a constant warm-up phase for the first 25\% of the total training steps before initiating the decay. For the WOMD dataset, to account for the increased dataset size, training is scaled to 8 GPUs achieving a convergence in 26 hours, trained for a total of 100 epochs (peak LR is $2.1 \times 10^{-3}$ using square root scaling law) with each training epoch taking an average of 14 min\footnote{Benchmarked for the last 40 epochs to account for kernel warmups}. 

Due to the higher dataset size of WOMD when compared with AV2, we also scaled the model parameters to 13.2 Million, resulting in R2P-L and R2P-M having 8.1 Million parameters. We have to acknowledge that if one follows the proper scaling law from \cite{baniodeh_scaling_2025}, then the parameter for the R2P-L should be anywhere between 18.0 and 22.0 Million but since we are limited by compute and constrained by the limited FLOPS in our possession, we choose 13.2 as x0.75 the total parameters when compared with the R2P-M model.

\section{Results and Discussion}

Our proposed framework, R2P-M, demonstrates competitive performance on the AV2 test splits, performing on par with and in certain geometric metrics exceeding established baselines like HPTR and ForecastMAE, while uniquely preserving transparent interpretability and using only ${1/4}^{th}$ of the map information. On the WOMD, our architecture successfully adapts to the dataset's challenging internal statistics and converges reliably. Notably, our WOMD validation scores are achieved cleanly, without relying on dense ensembling or complex post-processing heuristics, except for a simple temperature scaling of 0.5 applied to the final confidence scores during validation.

Evaluating standard geometric distance metrics, R2P-M achieves impressive $\text{minADE}_6$ and $\text{minFDE}_6$ scores across both benchmarks. As shown in \cref{tab:av2_test} (AV2), R2P yields a $\text{minFDE}_6$ of 1.33, actively outperforming both HPTR (1.43) and ForecastMAE (1.42), also in brier-minFDE (2.03). Similarly, on WOMD (\cref{tab:waymo_val}), our R2P-M achieves a $\text{minADE}_6$ of  0.54 and a $\text{minFDE}_6$ of 1.11, surpassing strong recent baselines such as EDA and MTR++. However, we observe slightly less competitive results in confidence-dependent metrics, specifically brier-minFDE on AV2 and mAP on WOMD. We attribute this slight discrepancy primarily to the uncalibrated nature of our mode confidence head. 

\begin{figure*}[t]
    \centering
    \begin{subfigure}[b]{0.32\textwidth}
        \fbox{\includegraphics[width=\textwidth]{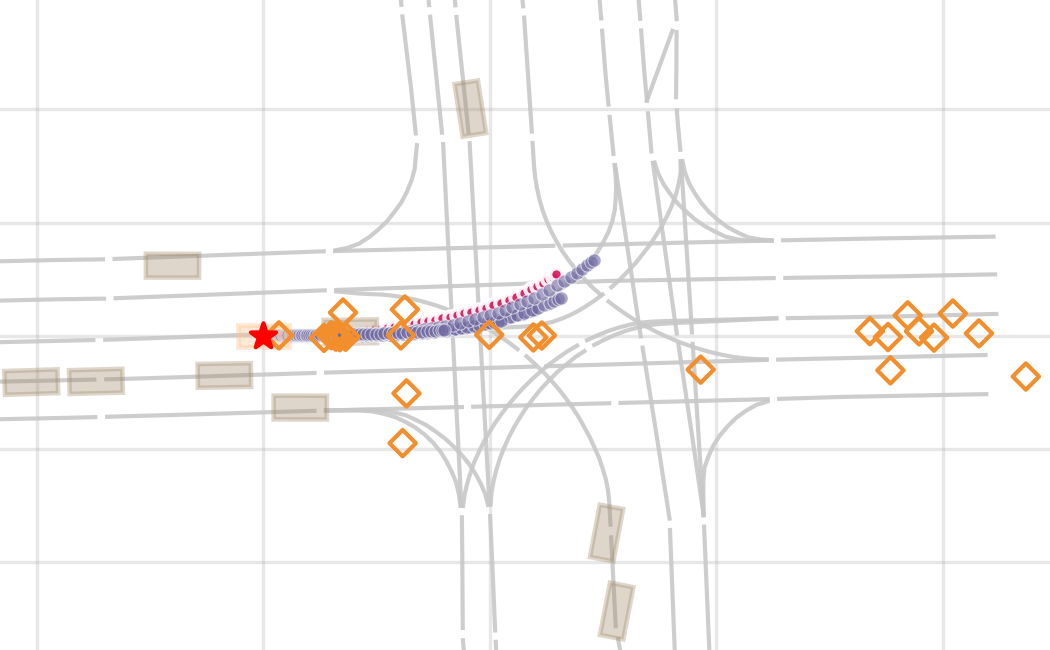}}
    \end{subfigure}
    \hfill
    \begin{subfigure}[b]{0.32\textwidth}
        \fbox{\includegraphics[width=\textwidth]{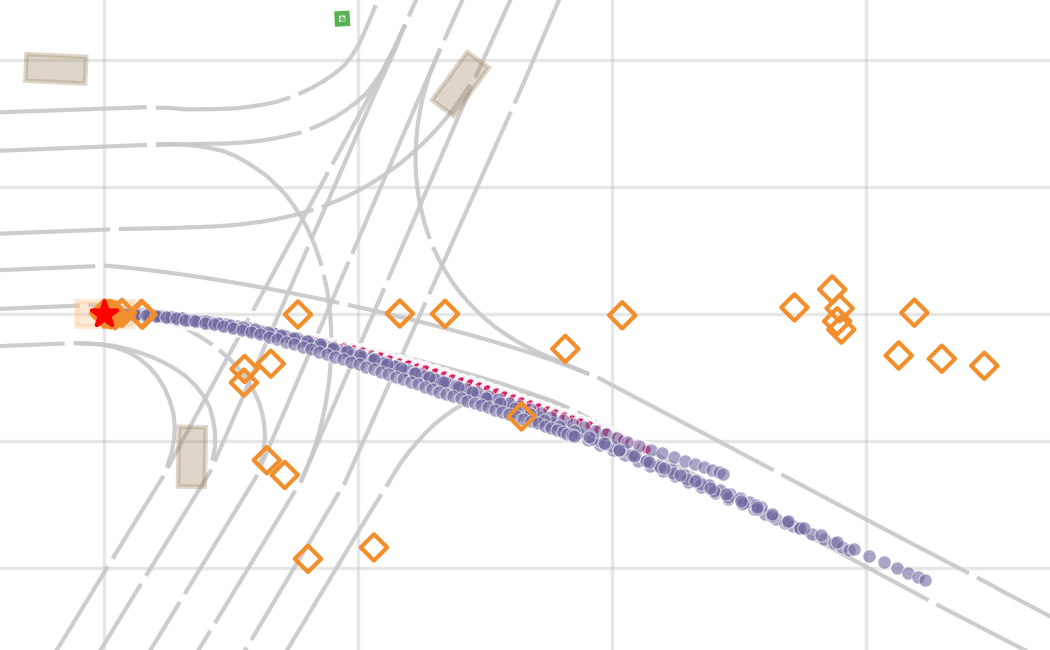}}
    \end{subfigure}
    \hfill
    \begin{subfigure}[b]{0.32\textwidth}
        \fbox{\includegraphics[width=\textwidth]{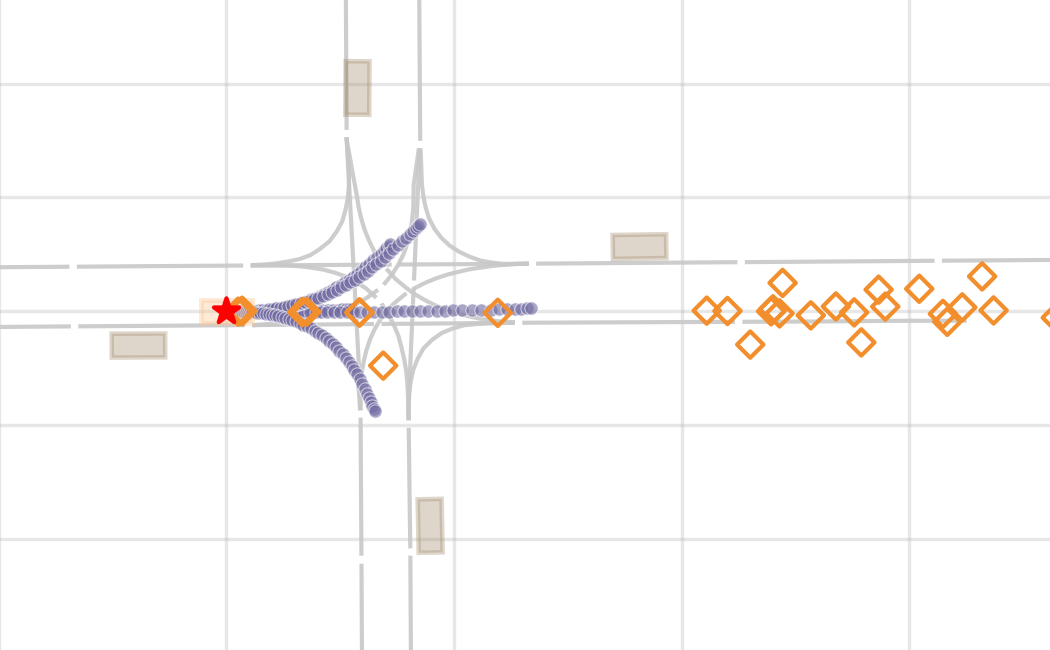}}
    \end{subfigure}

    \vspace{0.5em}

    \begin{subfigure}[b]{0.32\textwidth}
        \fbox{\includegraphics[width=\textwidth]{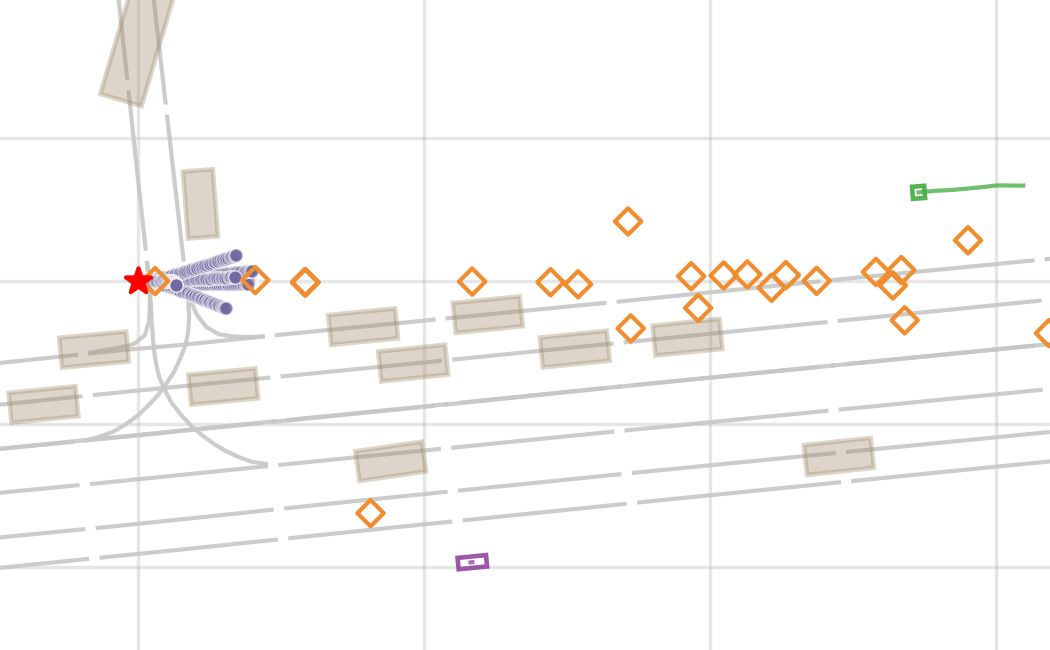}}
    \end{subfigure}
    \hfill
    \begin{subfigure}[b]{0.32\textwidth}
        \fbox{\includegraphics[width=\textwidth]{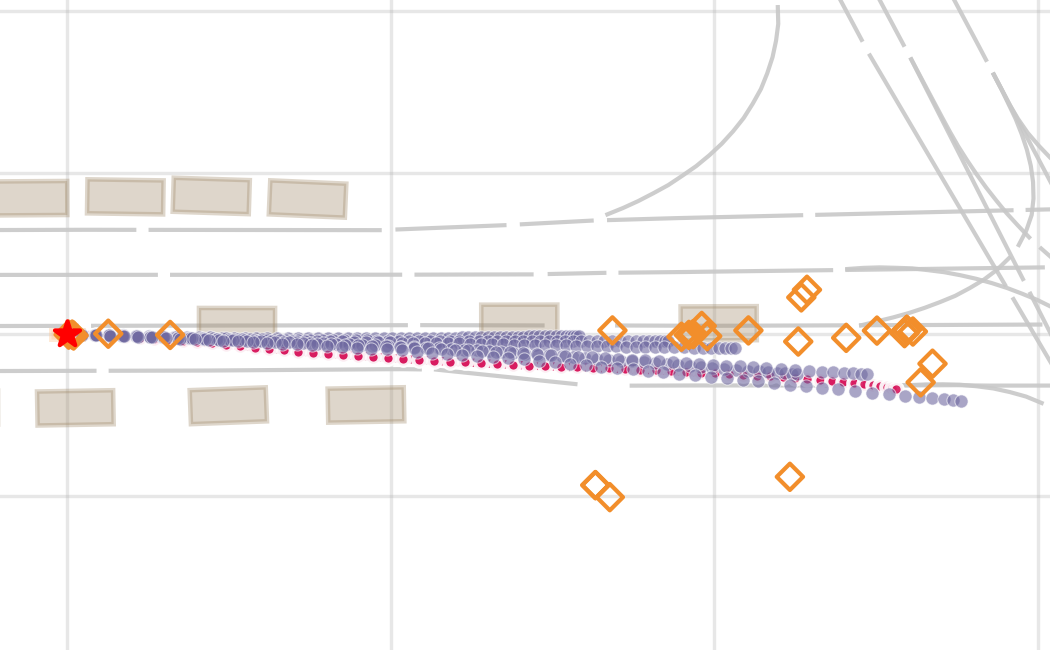}}
    \end{subfigure}
    \hfill
    \begin{subfigure}[b]{0.32\textwidth}
        \fbox{\includegraphics[width=\textwidth]{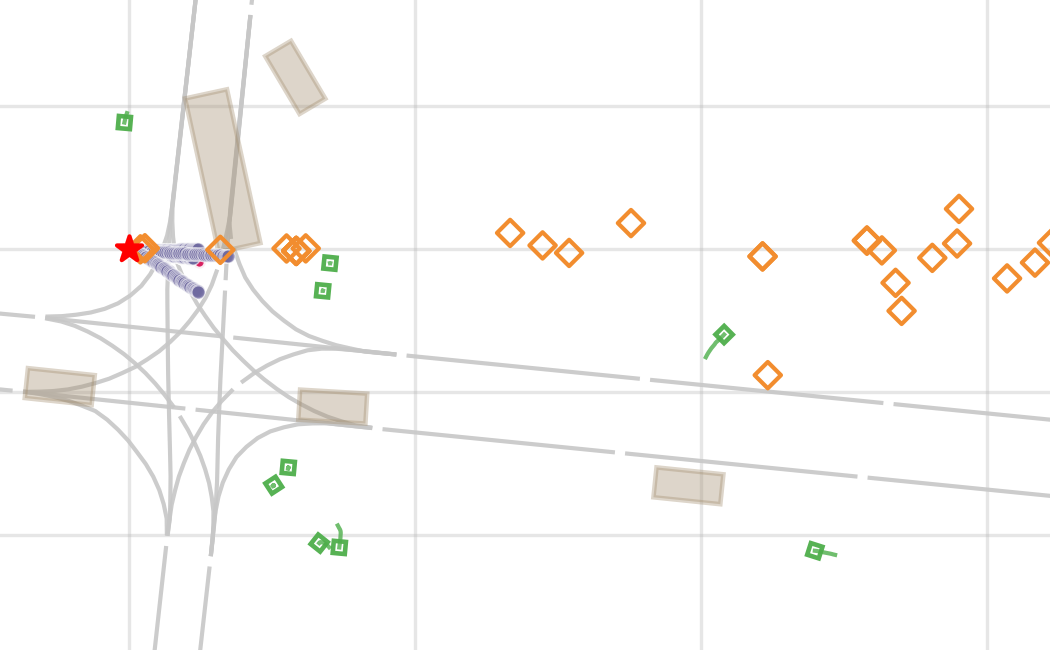}}
    \end{subfigure}
\caption{Qualitative forecasting results on the Argoverse 2 (AV2) dataset. The \textcolor{orange}{diamond markers} denote the spatial endpoints of 32 anchors retrieved by the Anchor Retrieval Layer, with the star marker indicating the last observed time step of the target agent. The network's refined multi-modal predictions are illustrated as \textcolor{predcolor}{solid trajectories}, while the ground truth path is highlighted in \textcolor{magenta}{pink}.}
    \label{fig:three_images}
\end{figure*}

Beyond benchmark comparisons, we investigated the structural properties of the motion bank. During training, models utilizing subsets constructed via random sampling yielded severely degraded metrics compared to those trained with mean clustering. Furthermore, under random sampling, the distance between the retrieved anchor endpoints and the ground-truth trajectory failed to converge meaningfully, indicating an inability of the network to reliably select appropriate priors. Conversely, models trained with clustered subsets similar to Fig.2 of \cite{vivekanandan_contrast_2025} successfully preserved multi-modal diversity and exhibited a steady, well-behaved decrease in endpoint error as can be seen from \cref{tab:bank_ablations}. This demonstrates that structured, cluster-based sampling provides a critical inductive bias for stable representation learning.

However, we acknowledge a current limitation regarding Vulnerable Road Users (VRUs), such as pedestrians and cyclists. Because the current motion bank is predominantly biased toward vehicular kinematics, its representation of VRUs remains relatively coarse. We hypothesize that employing a targeted, high-density clustering strategy specifically tailored to the long-tail distributions of VRUs will seamlessly resolve this bottleneck, presenting a highly promising direction for future work.

\paragraph{Interpretability of Context-Aware Retrieval.} \cref{fig:4 interp_explanation} illustrates how our orthogonally initialized queries dynamically distill representations from the available heterogeneous context. By inspecting the intermediate cross-attention weights, we observe a striking distribution: the macro-level routing mechanism assigns roughly 42\% of its attention mass to neighboring actors, compared to less than 8\% for the target agent's own history. Even within this marginal allocation to the target, attention is almost exclusively concentrated on the most recent four timesteps.

This finding yields a crucial architectural insight: rather than relying on its own extended past, the model predominantly infers its initial priors by contextualizing its immediate recent kinematics (the last four timesteps) with the rich, temporally-aggregated representations of neighboring actors. Furthermore, we observe that individual queries learn to capture unique social interactions. Instead of distributing their attention evenly across all vehicles in the scene, each query focuses sharply on a completely different neighboring actor to generate its distinct future prediction. Finally, the learnable ``null'' sink consistently absorbs approximately 50\% of the probability mass. This demonstrates that our gating strategy efficiently empowers queries to ignore irrelevant background elements, thereby maintaining the structural integrity and semantic purity of the retrieved anchor priors.

\section{Conclusion}

We presented Recall to Predict (R2P), an end-to-end motion forecasting framework that resolves the historical trade-off between predictive accuracy and multi-modal interpretability. By replacing opaque latent queries with a contrastively-trained motion bank, R2P grounds trajectory generation in a structured space of physically realizable driving behaviors. At its core, the framework leverages a novel Anchor Retrieval Layer with a Straight-Through Gumbel-Softmax estimator - to dynamically retrieve trajectory priors while maintaining continuous gradient flow. A subsequent DETR-style decoder, regularized by a latent diversity penalty and a Dual-Objective Endpoint Loss, strictly decouples global semantic intent from local geometric refinement. This ensures that continuous offset corrections are utilized solely for spatial calibration, preventing the network from lazily compensating for poor anchor selection.

Ultimately, R2P eliminates the ``black box'' nature of standard forecasting architectures. By exposing the exact motion primitives the model considers \textit{before} the decoding phase, our approach provides transparency into the network's intermediate reasoning. This natively constrained, interpretable initialization offers a compelling paradigm shift for autonomous driving, proving that state-of-the-art multi-modal forecasting can be achieved without sacrificing human-interpretable reasoning.
\section{Acknowledgement}
The research leading to these results is funded by the German Federal Ministry for Economic Affairs and Climate Action within the project “NXT GEN AI METHODS” and the authors gratefully acknowledge the Gauss Centre for Supercomputing e.V. (www.gauss-centre.eu) for funding this project by providing computing time on the GCS Supercomputer JUWELS \cite{JUWELS} at Jülich Supercomputing Centre (JSC)
{
    \small
    \bibliographystyle{ieeenat_fullname}
    \bibliography{main}
}
\newpage
\section{Supplementatal}

\subsection{Physical Grouped-Query Aggregation (PGQA)}
\label{sec:pgqa}

To process situations when $N_{q}$ $>$ $K$ and to maintain a constant computational overhead for the retrieved anchors which are to be used as queries in the decoder, we use Grouped Query Aggregation. This module mainly functions to reduce the unnecessary computational overhead when the number of generated anchors is more than the number of prediction queries which we require. To improve efficiency without sacrificing coverage, we compress the retrieved set into $K$ representative queries in the physical trajectory space, as many retrieved trajectories are spatially redundant and can be removed safely. We did a broad range of ablation on the $N_{q}$ as can be seen from \cref{sec:anchor_ablations} range to study its impact on the downstream metrics.

Let $N = N_q$ denote the number of retrieved anchors, $\mathbf{z}_i \in \mathbb{R}^{D}$ the latent token of the $i$-th anchor, and $\mathbf{Y}_i \in \mathbb{R}^{T_f \times 2}$ its associated physical trajectory. We first extract the final endpoint $\mathbf{p}_i = \mathbf{Y}_i^{(T_f)}$ for each anchor. To ensure spatial diversity, we select $K$ representative "seed" anchors using farthest point sampling (FPS) over the set of endpoints $\{\mathbf{p}_i\}_{i=1}^{N}$. The FPS algorithm is initialized with the highest-ranked retrieved anchor, and subsequent seeds are iteratively chosen to maximize the distance from the already selected set.

For each selected seed anchor $s_k$, we compute the squared Euclidean distance to every retrieved endpoint:
\begin{equation}
d_{k,i} = \left\| \mathbf{p}_i - \mathbf{p}_{s_k} \right\|_2^2.
\end{equation}
These physical distances are then converted into soft assignment weights via a temperature-scaled softmax:
\begin{equation}
a_{k,i} = \frac{\exp(-d_{k,i}/\tau_g)}{\sum_{j=1}^{N} \exp(-d_{k,j}/\tau_g)},
\end{equation}
where $\tau_g$ controls the sharpness of the assignment (distinct from the retrieval temperature $\tau$ in \cref{eqn:temp_anneal}). The grouped latent query corresponding to seed $s_k$ is constructed as a convex combination of all retrieved tokens:
\begin{equation}
\mathbf{q}_k = \sum_{i=1}^{N} a_{k,i}\,\mathbf{z}_i.
\end{equation}
Crucially, while the FPS seed selection is a discrete operation, the subsequent grouping remains fully differentiable. This allows gradient information to flow back to every retrieved anchor, ensuring that all $N_q$ proposals are shaped by the downstream prediction loss.

This design offers two key advantages. First, it significantly reduces the number of decoder queries while explicitly preserving the spatial diversity of the predictions. Second, it maintains a strict physical grounding: each grouped query remains tied to the exact trajectory of its selected representative medoid, such that $\hat{\mathbf{Y}}_k = \mathbf{Y}_{s_k}$.
\begin{figure}[t]
    \centering
    \includegraphics[width=0.6\columnwidth, keepaspectratio]{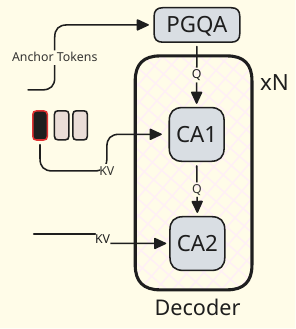}
    \caption{Optional Physical Grouped Query Aggregation (PGQA) layer. By default, the number of DETR queries equals the number of output
    modes ($K=6$). PGQA decouples these by aggregating multiple anchor-conditioned queries into each mode, enabling ablations on
    the number of trajectory anchors independently of $K$.}
    \label{fig:pgqa}
\end{figure}

Finally, to encourage softer assignments and prevent premature mode collapse, we compute the mean entropy of the assignment matrix $\mathbf{A_
{mat}} \in \mathbb{R}^{K \times N}$:
\begin{equation}
\mathcal{H}(\mathbf{A}) = -\frac{1}{K}\sum_{k=1}^{K}\sum_{i=1}^{N} a_{k,i}\log a_{k,i}.
\end{equation}
This entropy term can optionally be maximized during training as a regularization objective.

\begin{algorithm}[t]
\caption{Physical Anchor Grouped-Query Aggregation (per sample).}
\label{alg:pagqa}
\begin{algorithmic}[1]
\Require Latent tokens $\mathbf{Z}\!\in\!\mathbb{R}^{N\times D}$; trajectories $\mathbf{Y}\!\in\!\mathbb{R}^{N\times T_f\times 2}$; groups $K$; temperature $\tau_g$
\Ensure Grouped queries $\tilde{\mathbf{Z}}\!\in\!\mathbb{R}^{K\times D}$; medoid trajectories $\tilde{\mathbf{Y}}\!\in\!\mathbb{R}^{K\times T_f\times 2}$; assignment $\mathbf{A}\!\in\!\mathbb{R}^{K\times N}$
\State $\mathbf{e}_n \gets \mathbf{Y}_{n,T_f,:}$ \Comment{endpoint of each anchor}
\State $s_1 \gets 1$;\quad $d_n \gets \infty$ for all $n$
\For{$k = 1,\dots,K$}
  \State $d_n \gets \min\big(d_n,\,\lVert \mathbf{e}_n - \mathbf{e}_{s_k}\rVert_2^{\,2}\big)$
  \State $s_{k+1} \gets \arg\max_{n} d_n$ \Comment{FPS step; indices non-differentiable}
\EndFor
\State $\tilde{\mathbf{e}}_k \gets \mathbf{e}_{s_k}$;\quad $\tilde{\mathbf{Y}}_k \gets \mathbf{Y}_{s_k}$ \Comment{differentiable gather}
\State $\mathbf{A}_{k,n} \gets \mathrm{softmax}_n\!\big(-\lVert \mathbf{e}_n - \tilde{\mathbf{e}}_k \rVert_2^{\,2}\,/\,\tau_g\big)$
\State $\tilde{\mathbf{Z}}_{k,:} \gets \sum_{n} \mathbf{A}_{k,n}\,\mathbf{Z}_{n,:}$
\State \Return $\tilde{\mathbf{Z}},\,\tilde{\mathbf{Y}},\,\mathbf{A}$
\end{algorithmic}
\end{algorithm}

To maintain decoding efficiency, we compress the $N_q$ retrieved trajectory anchors into $K \ll N_q$ representative queries while preserving gradient flow to the original set. We select $K$ medoid anchors via Farthest Point Sampling (FPS) over their 2D endpoints, ensuring the groups cover distinct behavioral modes. Each of the $N$ anchors is then softly assigned to the $K$ medoids using a $\tau_g$-scaled softmax over the negative squared endpoint distances, producing an assignment matrix $\mathbf{A_{mat}} \in \mathbb{R}^{K \times N}$. The grouped query tokens are computed as $\tilde{\mathbf{Z}} = \mathbf{A_{mat}}\mathbf{Z}$, and the regression loss is applied against the exact medoid trajectories. Because the soft assignment is fully differentiable, every retrieved anchor—not just the $K$ medoids—receives gradient updates from the downstream trajectory loss.

\paragraph{Complexity and Hyperparameters.}
\textsc{PA-GQA} introduces minimal computational overhead, requiring only $\mathcal{O}(NK)$ distance evaluations between 2D endpoints and a single $K \times N \times D$ matrix multiplication. This cost is negligible compared to the self-attention mechanisms within the decoder. In our experiments, we perform ablations for a variety of $N_q$ retrieved anchors, but keep the $K=6$ consistently throughout the experiments. Ablations on these hyperparameters are detailed in Sec.~\ref{sec:ablations}.

\subsection{Ablation Studies on Motion Bank and Trajectory Selection}
\label{sec:ablations}

To validate our design choices for the trajectory selector and the motion bank, we conduct comprehensive ablations on the Argoverse 2 validation dataset.

\begin{table}[htbp]
\centering
\caption{Ablation study on motion bank bucket size and configuration.}
\label{tab:bank_ablations}
\resizebox{\columnwidth}{!}{%
\begin{tabular}{@{}lcccc@{}}
\toprule
\textbf{Bank Configuration} & \textbf{minADE$_{6}$} $\downarrow$ & \textbf{minFDE$_{6}$} $\downarrow$ & \textbf{MR} $\downarrow$ & \textbf{mAP} $\uparrow$ \\
\midrule
\multicolumn{5}{@{}l}{\textit{Impact of Motion Bank Size}} \\
100\% Baseline ($128 \times 64$) & 0.6741 & 1.2674 & 0.1473 & 0.2901 \\
44\% Subset & 0.6928 & 1.2967 & 0.1488 & 0.2741 \\
11\% Subset & 0.6938 & 1.3005 & 0.1536 & 0.2043 \\
\midrule
\multicolumn{5}{@{}l}{\textit{Impact of Bucket Size (Clusters $\times$ Elements)}} \\
$32 \times 128$ & \textbf{0.6676} & \textbf{1.2417} & \textbf{0.1371} & \textbf{0.2923} \\
$64 \times 64$ & 0.6698 & 1.2489 & 0.1426 & 0.2802 \\
$64 \times 128$ & 0.6819 & 1.2673 & 0.1458 & 0.2167 \\
9216 (Flat) & 0.6892 & 1.2936 & 0.1469 & 0.2269 \\
\bottomrule
\end{tabular}%
}
\end{table}

\paragraph{Motion Bank Capacity.}
\cref{tab:bank_ablations} demonstrates that reducing the motion bank size progressively degrades performance. Compressing the baseline to a 44\% subset drops mAP from 0.2901 to 0.2741, while extreme pruning to an 11\% subset causes mAP to collapse to 0.2043 alongside an increased miss rate (0.1536). This confirms that a sufficiently dense geometric prior is essential to capture the long tail of complex driving behaviors.

\paragraph{Bank Topology and Structure.}
Evaluating the retrieval bank's topology (Clusters $\times$ Elements) reveals that a $32 \times 128$ configuration optimizes all metrics (mAP 0.2923, minADE$_6$ 0.6676). While a $128 \times 64$ configuration remains competitive, topologies that retrieve excessive elements—such as $64 \times 128$ or a flat 9216 bank—cause severe mAP degradation (0.2167 and 0.2269, respectively). We hypothesize that retrieving too many anchors dilutes the attention weights, yielding noisy latent queries that lose their distinct multimodal characteristics.

\begin{table}[htbp]
\centering
\caption{Ablation study on PGQA and retrieval strategies. \textbf{ST} denotes the use of a discrete Straight-Through estimator for hard selection (as opposed to fully soft, differentiable retrieval). \textbf{Soft-R} denotes soft retrieval where the STE is disabled and the attention weights are directly used for making a weighted combination of the embeddings from the bank.}  
\label{tab:gqa_retrieval_ablations}
\resizebox{\columnwidth}{!}{%
\begin{tabular}{@{}lcccc@{}}
\toprule
\textbf{Bank Configuration} & \textbf{minADE$_{6}$} $\downarrow$ & \textbf{minFDE$_{6}$} $\downarrow$ & \textbf{MR} $\downarrow$ & \textbf{mAP} $\uparrow$ \\
\midrule
$32 \times 128$ (GQA, ST) & \textbf{0.6676} & \textbf{1.2417} & \textbf{0.1371} & \textbf{0.2923} \\
$32 \times 128$ (GQA, Soft-R) & 0.6973 & 1.2921 & 0.1519 & 0.2690 \\
\bottomrule
\end{tabular}%
}
\end{table}

\paragraph{Soft vs. Hard Retrieval (STE).}
We evaluated the impact of retrieval strategies by comparing a fully differentiable soft retrieval approach (Soft-R) against a discrete hard-selection bottleneck using a Straight-Through Estimator (STE). The hard-selection approach consistently outperformed soft retrieval across all metrics, achieving a lower minADE$_6$ (0.6676 vs. 0.6973) and higher mAP (0.2923 vs. 0.2690). We hypothesize that soft assignment dilutes the gradient flow by constructing anchors as a weighted combination of spatial trajectories. This blending degrades the learned representations and disrupts the one-to-one mapping required during the decoding phase. In contrast, the discrete selection enforced by the STE preserves distinct trajectory representations, validating our architectural design choice.

\subsection{Ablation on Map Polylines}
\label{sec:map_ablations}

\begin{table}[htbp]
\centering
\caption{Ablation on the number of map polylines ($N_{\text{map}}$) provided to the network.}
\label{tab:map_ablations}
\resizebox{\columnwidth}{!}{%
\begin{tabular}{@{}lcccc@{}}
\toprule
\textbf{Map Polylines ($N_{\text{map}}$)} & \textbf{minADE$_{6}$} $\downarrow$ & \textbf{minFDE$_{6}$} $\downarrow$ & \textbf{MR} $\downarrow$ & \textbf{mAP} $\uparrow$ \\
\midrule
128 & 0.7130 & 1.3490 & 0.1564 & 0.2480 \\
256 & \textbf{0.6690} & \textbf{1.2440} & \textbf{0.1381} & \textbf{0.2923} \\
512 & 0.7095 & 1.2990 & 0.1459 & 0.2570 \\
\bottomrule
\end{tabular}%
}
\end{table}

\paragraph{Impact of Map Context Size.}
To understand the model's reliance on static environmental geometry, we ablate the number of map polylines ($N_{\text{map}}$) processed by the network. As shown in Table~\ref{tab:map_ablations}, expanding the context from 128 to 256 polylines yields a significant improvement across all metrics, achieving the best performance (minADE 0.6690, mAP 0.2923). This highlights the critical role of comprehensive road topology in trajectory forecasting, as it allows the mechanism to query a wider receptive field. However, further doubling the capacity to 512 polylines degrades the performance metrics. We hypothesize that an excessively large context window introduces noise or irrelevant distant geometry, diluting the attention mechanism's ability to focus on the immediate, highly relevant drivable areas.

\subsection{Ablation on Network Gates}
\label{sec:gate_ablations}

\begin{table}[htbp]
\centering
\caption{Impact of selective gating mechanisms on trajectory metrics. Checkmarks ($\checkmark$) and crosses ($\times$) denote whether the specific context gate is active.}
\label{tab:gate_ablations}
\resizebox{\columnwidth}{!}{%
\begin{tabular}{@{}ccccccc@{}}
\toprule
\textbf{Target} & \textbf{Map} & \textbf{Other} & \textbf{minADE$_{6}$} $\downarrow$ & \textbf{minFDE$_{6}$} $\downarrow$ & \textbf{MR} $\downarrow$ & \textbf{mAP} $\uparrow$ \\
\midrule
$\times$     & $\times$     & $\times$     & 0.7292 & 1.3490 & 0.1550 & 0.2580 \\
$\checkmark$ & $\times$     & $\times$     & 0.6690 & 1.2554 & 0.1480 & 0.2923 \\
$\checkmark$ & $\times$     & $\checkmark$ & \textbf{0.6652} & \textbf{1.2481} & \textbf{0.1441} & \textbf{0.3077} \\
$\checkmark$ & $\checkmark$ & $\checkmark$ & 0.7247 & 1.3630 & 0.1647 & 0.2440 \\
\bottomrule
\end{tabular}%
}
\end{table}

\paragraph{Selective Context Routing.}
\cref{tab:gate_ablations} isolates the contribution of specific network gates in routing semantic context. Disabling all routing mechanisms yields a weak baseline (mAP 0.2580). Co-activating the Target and Other gates emerges as the strictly optimal configuration, achieving the best performance across all evaluated metrics, including the highest mAP (0.3077), lowest minADE$_6$ (0.6652), and lowest Miss Rate (0.1441). While enabling only the Target gate improves upon the baseline (mAP 0.2923), it falls short of this dual-gate approach. Conversely, activating all gates concurrently (Target, Map, and Other) severely degrades both spatial and probabilistic performance, dropping mAP to 0.2440 and increasing Miss Rate to 0.1647. This indicates that unconstrained context aggregation introduces noise, whereas selective gating effectively filters irrelevant semantic features.

\subsection{Ablation on Latent Anchors}
\label{sec:anchor_ablations}

\begin{table}[htbp]
\centering
\caption{Ablation on the number of latent anchors used for downstream forecasting.}
\label{tab:latent_ablations}
\resizebox{\columnwidth}{!}{%
\begin{tabular}{@{}lcccc@{}}
\toprule
\textbf{Latent Anchors} & \textbf{minADE$_{6}$} $\downarrow$ & \textbf{minFDE$_{6}$} $\downarrow$ & \textbf{MR} $\downarrow$ & \textbf{mAP} $\uparrow$ \\
\midrule
128 & 0.7030 & 1.3230 & 0.1520 & 0.2650 \\
64 & 0.7290 & 1.3650 & 0.1660 & 0.2590 \\
32 & 0.6900 & 1.2990 & 0.1460 & 0.2810 \\
16 & 0.6980 & 1.3040 & 0.1510 & 0.2790 \\
6 & \textbf{0.6741} & \textbf{1.2674} & \textbf{0.1473} & \textbf{0.2910} \\
\bottomrule
\end{tabular}%
}
\end{table}

\paragraph{Impact of the Number of Latent Anchors.}
Finally, we study the downstream impact of altering the number of bottleneck latent anchors. Table~\ref{tab:latent_ablations} reveals an inverse relationship between the number of latent anchors and overall accuracy. Configurations utilizing a high number of latent anchors (e.g., 128 or 64) experience noticeable degradation in minADE and mAP. As the bottleneck is compressed further, performance strictly improves. Constraining the network to exactly 6 latent anchors achieves the most competitive results across the board (minADE 0.6741, mAP 0.2910). This confirms that a tighter latent bottleneck effectively forces the network to converge on highly discriminative, high-quality multimodal trajectories.

\end{document}